\documentclass[letterpaper, 10 pt, conference]{ieeeconf}  

\IEEEoverridecommandlockouts                              

\usepackage[pdftex]{graphicx}
\usepackage{xcolor}
\usepackage{amsmath}
\usepackage{amssymb}
\usepackage{multirow}
\usepackage{cuted}
\usepackage{ulem}
\usepackage{siunitx}
\usepackage{threeparttable}
\usepackage{booktabs}
\usepackage{hyperref}
\usepackage[stable]{footmisc}

\title{\LARGE \bf
Fail-safe Flight of a Fully-Actuated Quadcopter\\ in a Single Motor Failure}

\author{Seung Jae Lee$^{1}$, Inkyu Jang$^{2}$ and H. Jin Kim$^{1}$
\thanks{$^{1}$Seung Jae Lee and H. Jin Kim are with the Department of Mechanical and Aerospace Engineering, School of Engineering and Automation and Systems Research Institute (ASRI), Seoul National University, Gwanak-gu, Seoul, Korea\ {\tt\small sjlazza@snu.ac.kr, hjinkim@snu.ac.kr}}%
\thanks{$^{2}$Inkyu Jang is with the Department of Mechanical and Aerospace Engineering, School of Engineering, Seoul National University, Gwanak-gu, Seoul, Korea\ {\tt\small leplusbon@snu.ac.kr}}%
}

\begin{document}

\maketitle
\thispagestyle{empty}
\pagestyle{empty}

\begin{abstract}

In this paper, we introduce a new quadcopter fail-safe flight solution that can perform the same four controllable degrees-of-freedom flight as a regular multirotor even when a single thruster fails.
The new solution employs a novel multirotor platform known as the $T^3$-Multirotor and utilizes a distinctive strategy of actively controlling the center of gravity position to restore the nominal flight performance.
A dedicated control structure is introduced, along with a detailed analysis of the dynamic characteristics of the platform that change during emergency flights.
Experimental results are provided to validate the feasibility of the proposed fail-safe flight strategy.
\end{abstract}

\section{INTRODUCTION}

Multi-rotor UAVs (hereafter called `multirotors') have various fuselage shapes depending on the number of thrusters, but they all share the same principle of controlling flight through four control inputs: roll, pitch, yaw torque and overall thrust \cite{drone_basics}.
Since the number of control inputs is four, multirotors generally require at least four thrusters, and severe loss of stability can occur if the number of available thrusters is reduced to less than four \cite{fail_quad_2}.
From this fact, we can see that the quadcopter configuration with four thrusters is the minimum requirement for a stable multirotor flight under general hardware configurations, and it is highly difficult to maintain a stable flight if one or more thrusters of the quadcopter fail.

\subsection{Related Works}
Although the degradation of the flight performance during actuator failure is unavoidable in conventional multirotors, several methods \cite{fail_quad_2}$\sim$\cite{fail_other_3}
have been proposed to address the failure of multirotor
flight.
For multirotors with more than four thrusters \cite{fail_other_1}$\sim$\cite{fail_other_3}, the platform's redundancy in the actuator is applied to recover full degree of multirotor flight.
However, for quadcopters, the number of actuators available in any motor failure scenario will always be less than four, so one or more controllable degrees of freedom (c-DOF) must be given up depending on the number of motors that have failed.
As a result, quadcopter-based fail-safe flight commonly gives up yaw motion instead of maintaining full control of three-dimensional translational motion \cite{fail_quad_2}$\sim$\cite{fail_quad_5} in a single motor failure to prevent collisions with foreign objects or terrain.
This approach prevents the crash and guarantees safe return/land, but causes continuous rotation of the fuselage with payloads or sensors.
Therefore, this approach could cause difficulties for multirotor to continue carrying out designated missions after motor failure.
Also, for multirotor flights utilizing on-board sensors, such as a camera for visual odometry, continuous camera rotation could drastically deteriorate the quality of the sensor/navigation data and result in unstable flight \cite{rot_blur}.

To solve the yaw rotation problem, the research of \cite{fail_aug_quad_1} adopted a tilt-rotor-type quadrotor platform with eight c-DOFs.
In this case, both translational and yaw motion can be controlled even with the quadrotor configuration.
However, this method has the disadvantage of increasing the weight and power consumption, since the servomotor must be installed on each and every thruster for fail-safe flight.

\subsection{Contributions and Outline}

\begin{figure}[t]
    \begin{center}
    \includegraphics[width=1\columnwidth]{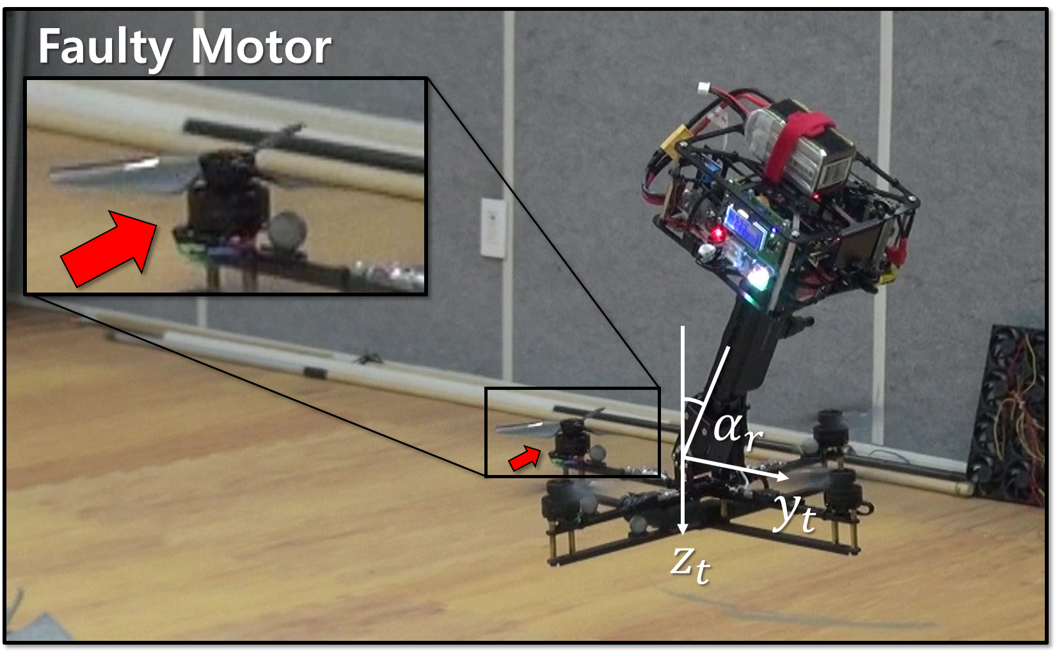}
    \end{center}
    \vspace{-0.5cm}
    \caption{The fail-safe flight of the $T^3$-Multirotor in a single-motor failure condition (red arrow indicating the faulty motor).}
    \vspace{-0.5cm}
    \label{gp:hardware}
\end{figure}

As a solution of the problems in previous fail-safe flight, in this paper, we introduce a new quadcopter fail-safe flight method utilizing the novel multirotor platform called the $T^3$-Multirotor (the platforms in Fig. \ref{gp:hardware}).\footnote{See \url{https://youtu.be/ePHoqFileuQ} for an experimental video.}
First introduced in \cite{icra_2018} and \cite{icra_2019}, the $T^3$-Multirotor platform is a fully-actuated quadcopter platform developed to overcome the dependency of fuselage attitude on translational motion.
By utilizing the unique mechanical features of the $T^3$-Multirotor, in this paper, we introduce a new strategy to restore the nominal flight performance by actively changing the platform's center of gravity position in a single motor failure scenario.

The remaining of the paper is organized as follows.
In Section II, the mechanism and dynamics of the $T^3$-Multirotor are introduced.
Section III introduces a fail-safe flight strategy, followed by an introduction to the fail-safe controller in Section IV.
In section V, the actual experimental results are provided with detailed analysis to validate the feasibility of the proposed method.

\section{MECHANISM \& DYNAMICS}

In this section, we briefly describe the mechanism of the $T^3$-Multirotor and introduce the equations of motion (EoMs).

\subsection{Mechanism}

\begin{figure}[t]
    \begin{center}
    \includegraphics[width=8.55cm]{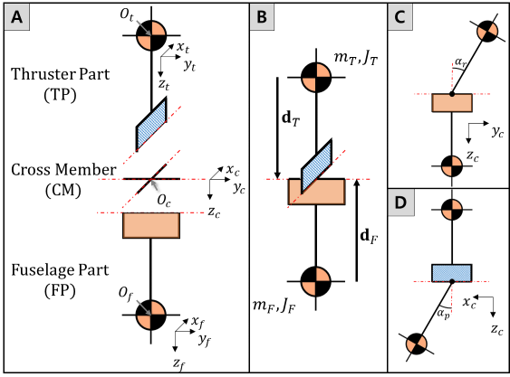}
    \end{center}
    \vspace{-0.38cm}
    \caption{Schematic of the $T^3$-Multirotor. The platform consists of three major parts, TP, FP, and CM, where the relative roll ($\alpha_r$) and pitch ($\alpha_p$) attitude between TP and FP can be actively modified.}
    \vspace{-0.6cm}
    \label{gp:CoG_figure}
\end{figure}

Fig. \ref{gp:CoG_figure} shows the schematic of the $T^3$-Multirotor.
As shown in Fig. \ref{gp:CoG_figure}-A, the platform consists of three major parts: Thruster Part (TP), Fuselage Part (FP), and Cross Member (CM).

The TP consists only of a frame with four arms, four propeller thrusters, and a single IMU sensor.
With thrusters, the TP can generate attitude control torques and overall thrust on the same principle as a regular quadcopter.
Meanwhile, the FP consists of the remaining components other than those mounted on the TP (e.g. battery, mission computer, auxiliary sensors).
Since there are no thrusters on FP, it cannot generate thrust on its own.
Instead, the FP is connected to the TP via the universal joint mechanism to receive the force required for flight.

The TP and FP are connected to each of the two rotation axes of the CM, which is a cross-shaped rigid body with two orthogonal rotational axes (refer Fig. \ref{gp:CoG_figure}-A and B), whereby the TP and FP have degrees of freedom in roll and pitch rotation for the CM, respectively (refer Fig. \ref{gp:CoG_figure}-C and D).
Then, the servomotors are attached to the roll and pitch axes of CM to actively control the relative attitude $\alpha_{\{r,p\}}\in\mathbb{R}$ (refer Fig. \ref{gp:CoG_figure} for the definition of $\alpha_*$).
This feature allows the FP to take an arbitrary attitude independent of the TP while the TP performs attitude control for translational motion control as in the conventional multirotor.

\begin{figure}[t]
    \begin{center}
    \includegraphics[width=8.1cm]{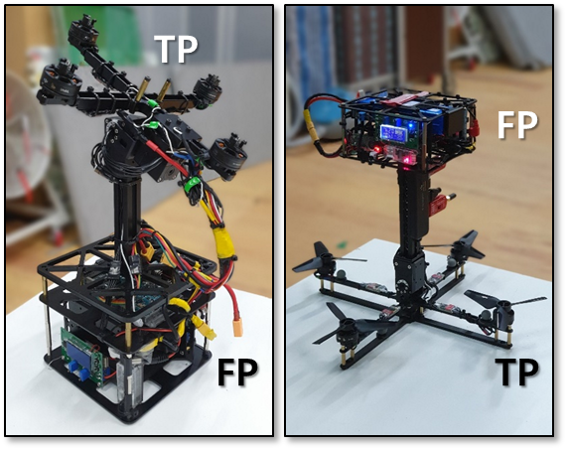}
    \end{center}
    \vspace{-0.5cm}
    \caption{Examples of $T^3$-Multirotor structures that can vary depending on mission objectives. $T^3$-Upright (Left): Case where TP is on top of FP. $T^3$-Inverted (Right): Case where FP is on top of TP.}
    \vspace{-0.5cm}
    \label{gp:t3_inverted}
\end{figure}

As shown in Fig. \ref{gp:t3_inverted}, the structure of the $T^3$-Multirotor is not limited to the shape shown in Fig. \ref{gp:CoG_figure} and may vary depending on the purpose of the mission.
For example, if the platform needs to interact downward (e.g. cargo transportation \cite{icra_2019}, ground observation), a structure with FP at the bottom is desirable.
On the contrary, in a mission requiring upward interaction (e.g. in-air object surveillance, interaction with the ceiling), a structure in which the TP is located at the bottom is preferable.
Despite differences in shape, however, the two structures have almost identical operating principles and motion dynamics, thus we can control the system in the same manner.

\subsection{Platform Dynamics}

The equations governing the motion of a $T^3$-Multirotor are already known from \cite{icra_2018} and \cite{icra_2019}.
Therefore, in this paper, we take the final form of the dynamic equations without the derivation process but explain the physical meaning of the terms that make up the equation.\footnote{See \url{https://www.sjlazza.com/post/ra-l-2020} for the complete process of the dynamics derivation process of the $T^3$-Multirotor.}

\subsubsection{Translational EoMs}
From the aforementioned previous researches, we can express the translational EoMs of the platform as
\begin{equation}
    \label{eq:trnas_EoM_simplified}
    \left\{
    \begin{array}{lr}
    \ddot{\mathbf{X}}_T\approx\frac{1}{M}\mathbf{R}(\mathbf{q}_T)\mathbf{F}_T+\mathbf{g}\\[3pt]
    \ddot{\mathbf{X}}_F\approx\ddot{\mathbf{X}}_T
    \end{array}
    \right.,
\end{equation}
where the subscripts $(*)_T$ and $(*)_F$ represent TP and FP, $\mathbf{X}_*=[x_{1,*}\ x_{2,*}\ x_{3,*}]^\intercal\in\mathbb{R}^{3\times1}$ is the position vector, $M$ is the overall mass of the platform, $\mathbf{q}_*=[q_{1,*}\ q_{2,*}\ q_{3,*}]^\intercal\in\mathbb{R}^{3\times1}$ is the Roll-Pitch-Yaw Euler attitude vector, $\mathbf{R}(\mathbf{q}_*)=\mathbf{R}_y(q_{3,*})\mathbf{R}_p(q_{2,*})\mathbf{R}_r(q_{1,*})\in SO(3)$ is the rotation matrix from the body frame to the Earth-fixed frame, $\mathbf{F}_T=[0\ 0\ -F_T]^\intercal\in\mathbb{R}^{3\times1}$ is the thrust force vector generated by the TP defined in the body frame of the TP, and $\mathbf{g}=[0\ 0\ g]^\intercal\in\mathbb{R}^{3\times1}$ is the gravitational acceleration vector.

From Eq. (\ref{eq:trnas_EoM_simplified}), we can see that the translational motion of the $T^3$-Multirotor is determined only by the total thrust $\mathbf{F}_T$ and TP attitude $\mathbf{q}_T$ as in the regular multirotor. 
Thus, we define a new vector $\mathbf{I}_X=[\mathbf{q}_T\ F_T]^\intercal\in\mathbb{R}^{4\times1}$ and manipulate this vector for controlling the translational motion.

\subsubsection{Rotational EoMs}
Also from the previous researches, we can represent the rotational EoM of the platform as
\begin{equation}
    \label{eq:rot_EoM_simplified}
    \small
    \left\{
    \begin{array}{lr}
    \mathbf{J}_T\ddot{\mathbf{q}}_T\approx\mathbf{T}_T+\mathbf{T}_{CT}=\mathbf{T}_O\\[3pt]
    \mathbf{J}_F\ddot{\mathbf{q}}_F\approx\left(\frac{m_F}{M}\right)\left(\mathbf{d}_F\times\left(\mathbf{R}_r(\alpha_r)\mathbf{R}_p(-\alpha_p)\mathbf{F}_T\right)\right)+\mathbf{T}_{CF}
    \end{array}
    \right.,
    \normalsize
\end{equation}
where $\mathbf{J}_*=diag(J_{1,*}, J_{2,*}, J_{3,*})\in\mathbb{R}^{3\times3}$ is the moment of inertia (MoI), $\mathbf{T}_T=[\tau_{1,T}\ \tau_{2,T}\ \tau_{3,T}]^\intercal\in\mathbb{R}^{3\times1}$ is the torque vector generated by the array of thrusters attached to the TP, $\mathbf{T}_{C*}\in\mathbb{R}^{3\times1}$ is the interactive torque vector acting from the CM to part $(*)$ defined in the body frame of the part $(*)$, and $\mathbf{T}_O=[\tau_{1,O}\ \tau_{2,O}\ \tau_{3,O}]^\intercal$ is the overall torque vector which determines the attitudinal motion of the TP.
The symbol $m_*$ represents the mass of the corresponding part and $\mathbf{d}_*=[0\ 0\ d_*]^\intercal\in\mathbb{R}^{3\times1}$ is the distance vector from the $O_*$ to the $O_c$ defined in the body coordinate of the object (*) (refer Fig. \ref{gp:CoG_figure}). 
For $\alpha_*$, the approximate relationships $\alpha_r\approx q_{1,T}-q_{1,F}$ and $\alpha_p\approx q_{2,T}-q_{2,F}$ generally hold under normal flight conditions.


The interaction torque $\mathbf{T}_{C(*)}$ consists of the following subterms:
\begin{equation}
    \label{eq:T_CT_T_CF_detail}
    \mathbf{T}_{C(*)}=\mathbf{T}_{s,C(*)}+\mathbf{T}_{f,C(*)},
\end{equation}
where $\mathbf{T}_{s,CT}=[\tau_{rs}\ 0\ 0]^\intercal$ and $\mathbf{T}_{s,CF}=[0\ \tau_{ps}\ 0]^\intercal$ are servo-generated torques, $\mathbf{T}_{f,CT}=[0\ \tau_{p,CT}\ \tau_{y,CT}]^\intercal$ and $\mathbf{T}_{f,CF}=[\tau_{r,CF}\ 0\ \tau_{y,CF}]^\intercal$ are the torques transferred by the CM structure.
In the idle relative attitude condition ($\alpha_{\{r,p\}}=0$), it is known that the interaction torques $\mathbf{T}^{idle}_{CT}$, $\mathbf{T}^{idle}_{CF}$ are as follows \cite{icra_2018,icra_2019}.
\begin{equation}
    \label{eq:T_CT_T_CF_nominal}
    \mathbf{T}^{idle}_{CT}=
    \begin{bmatrix}
    \tau_{rs}\\
    -\tau_{ps}\\
    -\frac{J_{3,F}}{J_{3,T}+J_{3,F}}\tau_{3,T}
    \end{bmatrix}, \mathbf{T}^{idle}_{CF}=-\mathbf{T}^{idle}_{CT}
\end{equation}
And the following relationship is generally established unless the relative attitude $\alpha_{\{r,p\}}$ are huge.
\begin{equation}
    \label{eq:TCTF}
    \mathbf{T}_{C*}\approx\mathbf{T}^{idle}_{C*}
\end{equation}

\section{FAIL-SAFE FLIGHT}

In this section, we introduce a control strategy that utilizes the unique feature of the $T^3$-Multirotor to provide full control of the multirotor even under single motor failure condition.
Then, we analyze the EoM of the system during fail-safe flight to understand the altered motion behavior.
The method proposed throughout this paper basically responds to single motor failure, but a methodology for dealing with failures of two motors under special circumstances is also introduced.

\subsection{Fail-safe Flight - Strategy}

As well known, the translational motion of the conventional multirotor is determined by the overall thrust force and the fuselage attitude.
The attitude and thrust of the multirotor are generally controlled by $\mathbf{u}\in\mathbb{R}^{4\times1}$ value \cite{drone_basics}, which is generated by combining four motor thrusts $\mathbf{c}_0\in\mathbb{R}^{4\times1}$ as
\begin{equation}
    \label{eq:u_conven}
    \mathbf{u}=
    \begin{bmatrix}
    \mathbf{T}_T\\
    F_T
    \end{bmatrix}
    =
    \mathbf{A}_0\mathbf{c}_0=
    \begin{bmatrix}
    0   &   l   &   0    &  -l\\
    l   &   0   &   -l  &   0\\
    \frac{b}{k}&-\frac{b}{k}&\frac{b}{k}&-\frac{b}{k}\\
    1   &   1   &  1   &   1
    \end{bmatrix}
    \mathbf{c}_0,\ \mathbf{c}_0=
    \begin{bmatrix}
    F_1\\
    F_2\\
    F_3\\
    F_4
    \end{bmatrix}.
\end{equation}
The symbol $l\in\mathbb{R}$ represents the length of the arm frame, $b,k\in\mathbb{R}$ represent the torque and force coefficient of the thruster, and $F_i\in\mathbb{R}\geq0$ represents the thrust force generated by Motor $i$.
Here, we can see that there is no redundancy of $\mathbf{c}_0$ in generating the $\mathbf{u}$ signal with quadcopter configuration.
Accordingly, it is understood that the arbitrary required $\mathbf{u}$ value cannot be generated if any of the $F_i$ signals constituting $\mathbf{c}_0$ are out of control.


As with the conventional multirotor, we can see from Eq. (\ref{eq:trnas_EoM_simplified}) that the $T^3$-Multirotor determines the translational motion in four terms, $\mathbf{q}_T$ and $F_T$. 
Therefore, the $T^3$-Multirotor also cannot achieve fully stable flight if the number of actuators is less than four. 
However, unlike conventional quadcopters, the $T^3$-Multirotor has additional actuators other than thrusters, which are two servomotors that control relative roll and pitch attitude $\alpha_{\{r,p\}}$ between TP and FP.
Therefore, if we can take advantage of two servomotors to generate some of the components of $\mathbf{u}$ independently of $\mathbf{c}_0$, then we can use this feature to obtain additional redundancy in control and utilize in fail-safe flight.

\subsubsection{Control redundancy}


\begin{figure}[t]
    \begin{center}
    \includegraphics[width=3.6cm]{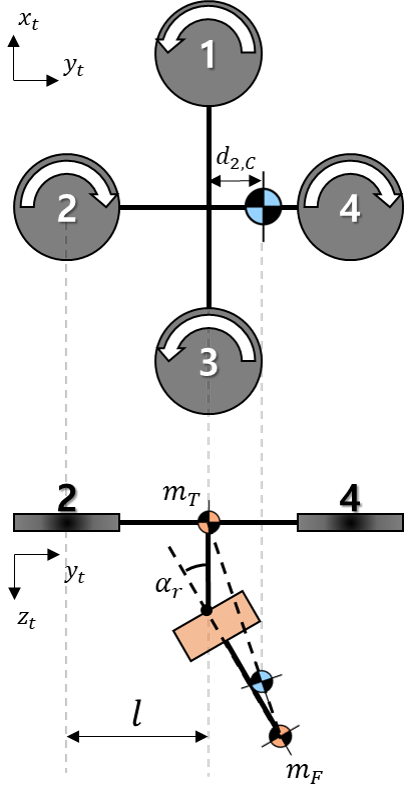}
    \end{center}
    \vspace{-0.5cm}
    \caption{The $d_{c,\{x,y\}}$, the overall CoG position of the platform w.r.t. the body coordinate system of the TP, can be changed by manipulating the relative attitude $\alpha_{\{r,p\}}$.}
    \vspace{-0.5cm}
    \label{gp:CoG_shift}
\end{figure}

When controlling the attitude of the FP via the servomechanism, the center of gravity (CoG) position with respect to the body coordinate of the TP changes with $\alpha_{\{r,p\}}$.
Fig. \ref{gp:CoG_shift} shows an example of the y-directional CoG position being changed in accordance of the relative roll attitude $\alpha_r$.
In the same manner, we can also change the CoG position in the x-direction with relative pitch attitude $\alpha_p$.
The position of the altered CoG with respect to the TP frame is represented by $\mathbf{d}_{C}=[d_{1,C}\ d_{2,C}\ d_{3,C}]^\intercal\in\mathbb{R}^{3\times1}$, and the relationship between the relative attitudes and the CoG position is as follows.
\begin{equation}
    \label{eq:X_CoG}
    \begin{split}
    &\mathbf{d}_{C}\left(\alpha_r,\alpha_p\right)=\\
    &\frac{1}{M}\left(m_T\mathbf{z}_{3\times1}+m_F\left(\mathbf{d}_T+\mathbf{R}_r(-\alpha_r)\mathbf{R}_p(\alpha_p)(-\mathbf{d}_F)\right)\right)
    \end{split}
\end{equation}
The symbol $\mathbf{z}_{m\times n}\in\mathbb{R}^{m\times n}$ represents the $m$-by-$n$-dimensional zero vector.
Then, we can find the relationship between relative attitude $\alpha_{\{r,p\}}$ and $d_{c,\{x,y\}}$ derived from Eq. (\ref{eq:X_CoG}) as follows.
\begin{equation}
    \label{eq:X_CoG_to_rel_att}
    \left\{
    \begin{array}{lr}
    \alpha_p=\arcsin{\left(\frac{M}{m_F}\frac{d_{1,C}}{d_F}\right)}\\
    \alpha_r=\arcsin{\left(\frac{1}{\cos{\left(\alpha_p\right)}}\frac{M}{m_F}\frac{d_{2,C}}{d_F}\right)}
    \end{array}
    \right.
\end{equation}
From Eq. (\ref{eq:X_CoG_to_rel_att}), we can calculate the required servomotor angles to relocate the CoG position on the ($x_t-y_t$) plane of the TP body coordinate to the desired location.

In the situation where the position of CoG is changing, the relationship between $\mathbf{c}_0$ and $\mathbf{u}$ is no longer static and does not follow the rule of Eq. (\ref{eq:u_conven}) anymore.
Instead, a new relationship that reflects the change in the position of CoG is established as follows
\begin{equation}
    \label{eq:T_T_updated}
    \small
    \left\{
    \begin{array}{lr}
    \tau^{CoG}_{1,T}=\left(l+d_{2,C}\right)F_2+d_{2,C}\left(F_1+F_3\right)-\left(l-d_{2,C}\right)F_4\\[3pt]
    \tau^{CoG}_{2,T}=\left(l-d_{1,C}\right)F_1-d_{1,C}\left(F_2+F_4\right)-\left(l+d_{1,C}\right)F_3\\[3pt]
    \tau^{CoG}_{3,T}=\frac{b}{k}\left(F_1-F_2+F_3-F_4\right)
    \end{array}
    \right.,
\end{equation}
where
\begin{equation}
    \label{eq:T_T_updated2}
    \mathbf{T}^{CoG}_T=
    \begin{bmatrix}
    \tau^{CoG}_{1,T}\\[3pt]
    \tau^{CoG}_{2,T}\\[3pt]
    \tau^{CoG}_{3,T}
    \end{bmatrix}
    =\mathbf{T}_T+
    \begin{bmatrix}
    d_{2,C}\\
    -d_{1,C}\\
    0
    \end{bmatrix}F_T.
\end{equation}
From Equations (\ref{eq:T_T_updated}) and (\ref{eq:T_T_updated2}), we can find that not only a series of $F_i$ values but also $d_{\{1,2\},C}$ values can be utilized to generate roll and pitch attitude control torques.
Based on this idea, we can change the way of generating the control input $\mathbf{u}$ from Eq. (\ref{eq:u_conven}) to
\begin{equation}
    \label{eq:u_generation}
    \small
    \mathbf{u}=
    \begin{bmatrix}
    \mathbf{T}_O\\
    F_T
    \end{bmatrix}
    =
    \begin{bmatrix}
    \mathbf{T}^{CoG}_T+\mathbf{\Delta}_d\\
    F_T
    \end{bmatrix}
    = \mathbf{A}_{aug}(F_T)\mathbf{c}_{aug}+
    \begin{bmatrix}
    \mathbf{\Delta}_d\\
    0
    \end{bmatrix},
\end{equation}
where $\mathbf{A}_{aug}(F_T)\in\mathbb{R}^{4\times6}$ is the new relationship between $\mathbf{c}_{aug}$ and $\mathbf{u}$, which is expressed as follows:
\begin{equation}
\label{eq:Aaug}
    \mathbf{A}_{aug}(F_T)=
    \begin{bmatrix}
    0   &   l   &   0   &   -l  &   0   &   F_T\\
    l   &   0   &   -l  &   0   &   -F_T&   0\\
    \frac{b}{k}&-\frac{b}{k}&\frac{b}{k}&-\frac{b}{k}&0&0\\
    1   &   1   &   1   &   1   &   0   &   0
    \end{bmatrix}.
\end{equation}
The symbol $\mathbf{c}_{aug}=[\mathbf{c_0}^\intercal \mathbf{d}^\intercal]^\intercal\in\mathbb{R}^{6\times1}$ is an augmented input vector, and $\mathbf{d}=[d_{1,C}\ d_{2,C}]^\intercal\in\mathbb{R}^{2\times1}$ is the altered $x$ and $y$ position of the CoG.
The term $\mathbf{\Delta}_d\in\mathbb{R}^{3\times1}$ is the gap between $\mathbf{T}_O$ and $\mathbf{T}_T^{CoG}$, which is an inertial term that occurs during the control of $\mathbf{d}$. 

To understand $\mathbf{\Delta}_d$, we need to revisit the rotational dynamics of the $T^3$-Multirotor.
By applying equations (\ref{eq:T_CT_T_CF_nominal}) and (\ref{eq:TCTF}) to (\ref{eq:rot_EoM_simplified}), we get the following relationship:
\begin{equation}
    \label{eq:rot_EoM_final}
    \begin{split}
    \mathbf{J}_T\ddot{\mathbf{q}}_T&=\mathbf{T}_O\\
    &=\mathbf{T}_T
    +\frac{m_Fd_F}{M}
    \begin{bmatrix}
    s(\alpha_r)c(\alpha_p)\\
    -s(\alpha_p)\\
    0
    \end{bmatrix}
    F_T-\mathbf{J}_F\ddot{\mathbf{q}}_F.
    \end{split}
\end{equation}
Next, by applying Eqs. (\ref{eq:X_CoG_to_rel_att}) and (\ref{eq:T_T_updated2}) to (\ref{eq:rot_EoM_final}), we can obtain the following equation:
\begin{equation}
    \label{eq:rot_EoM_finall}
    \mathbf{T}_O=\mathbf{T}_T
    +F_T
    \begin{bmatrix}
    d_{2,C}\\
    -d_{1,C}\\
    0
    \end{bmatrix}
    -\mathbf{J}_F\ddot{\mathbf{q}}_F=\mathbf{T}^{CoG}_T-\mathbf{J}_F\ddot{\mathbf{q}}_F.
\end{equation}
By comparing Eqs. (\ref{eq:u_generation}) and (\ref{eq:rot_EoM_finall}), we can see that $\mathbf{\Delta}_d=-\mathbf{J}_F\ddot{\mathbf{q}}_F$. 

Of the components of $\mathbf{T}_O$, $\mathbf{\Delta}_d$ is the only term that cannot be directly controlled.
Therefore we first consider $\mathbf{T}_T^{CoG}$ as the only control input for $\mathbf{q}_T$ control by ignoring the effects of $\mathbf{\Delta}_d$, but we will discuss the effects of $\mathbf{\Delta}_d$ during flight later by discovering the relationship between $\mathbf{T}_{O,d}$ and $\mathbf{T}_O$. 
The symbol $(*)_d$ represents the desired value.
Then, from  Eq. (\ref{eq:u_generation}), we can see that $\mathbf{c}_{aug}$ has two additional input redundancies in generating $\mathbf{u}$ signal. 
Based on this, we can now discuss the fail-safe flight strategy.

\subsubsection{Fail-safe flight strategy}
In the case of a single motor failure, our strategy is to modify Eq. (\ref{eq:u_generation}) to exclude the contribution of the faulty motor in the generation of $\mathbf{u}$, and utilize one of the components of $\mathbf{d}$ instead. 
During the process, remaining component of $\mathbf{d}$ is fixed to zero throughout the flight.
The strategy can be expressed as the following equation:
\begin{equation}
    \label{eq:u_generation_failsafe}
    \mathbf{u}=
    \begin{bmatrix}
    \mathbf{T}_O\\
    F_T
    \end{bmatrix}
    \approx\mathbf{A}_k(F_T)\mathbf{c}_k,
\end{equation}
where 
\begin{equation}
    \label{eq:AkCk}
    \left\{
    \begin{array}{lr}
    \mathbf{A}_k(F_T)=\mathbf{A}_{aug}(F_T)\mathbf{E}_k\\
    \mathbf{c}_k=\mathbf{E}_k^\intercal\mathbf{c}_{aug}
    \end{array}
    \right..
\end{equation}
The symbol $k=1,2,3,4$ represents the faulty motor, and $\mathbf{E}_k\in\mathbb{R}^{6\times4}$ represents the exclusion matrix defined as
\begin{equation}
    \label{eq:Ek}
    \mathbf{E}_k=
    \left\{
    \begin{array}{lr}
    \mathbf{I}_{\{k,5\}}\ (k=2,4)\\
    \mathbf{I}_{\{k,6\}}\ (k=1,3)
    \end{array}
    \right..
\end{equation}
The symbol $\mathbf{I}_{\{a,b\}}$ is a matrix in which two zero row vectors $\mathbf{z}_{1\times4}$ are added to the identity matrix $\mathbf{I}_4\in\mathbb{R}^{4\times4}$, resulting in all zero components of rows $a$ and $b$ of $\mathbf{I}_{\{a,b\}}$. 
In Eq. (\ref{eq:Ek}) we see that $\mathbf{E}_k$ is divided into two modes depending on the faulty motor $k$. 
We define this as \textit{Mode} $i$ ($i=1,2$), where \textit{Mode} 1 is the case when $k=2,4$ and \textit{Mode} 2 is the case when $k=1,3$.

The role of $\mathbf{E}_k$ is to exclude the faulty motor and one of the components of $\mathbf{d}$ from the Eq. (\ref{eq:u_generation}).
This process allows $\mathbf{A}_k(F_T)$ of Eq. (\ref{eq:u_generation_failsafe}) to have $rank(\mathbf{A}_k)=4$, restoring the system to a full rank even after the motor failure.
For example, in the event when Motor 2 fails, the \textit{Mode} 1 fail-safe flight scenario is activated, resulting in an $\mathbf{E}_2$ matrix as follows.
\begin{equation}
\label{eq:e2}
    \mathbf{E}_2^\intercal=
    \begin{bmatrix}
    1&0&0&0&0&0\\
    0&0&1&0&0&0\\
    0&0&0&1&0&0\\
    0&0&0&0&0&1
    \end{bmatrix}
\end{equation}
Then, Eq. (\ref{eq:u_generation_failsafe}) becomes
\begin{equation}
\label{eq:uM2}
    \mathbf{u}\approx\mathbf{A}_2(F_T)\mathbf{c}_2=
    \begin{bmatrix}
    0&0&-l&F_T\\
    l&-l&0&0\\
    b/k&b/k&-b/k&0\\
    1&1&1&0
    \end{bmatrix}
    \begin{bmatrix}
    F_1\\
    F_3\\
    F_4\\
    d_{2,C}
    \end{bmatrix}.
\end{equation}
From Eq. (\ref{eq:uM2}), we can calculate the required $\mathbf{c}_2$ to generate  $\mathbf{u}_d$ as
\begin{equation}
    \label{eq:c2foru}
    \mathbf{c}_{2,d}=\mathbf{A}_2^{-1}(F_T)\mathbf{u}_d=
    \begin{bmatrix}
    \frac{1}{2l}\tau_{2,O,d}+\frac{k}{4b}\tau_{3,O,d}+\frac{1}{4}F_{T,d}\\[3pt]
    -\frac{1}{2l}\tau_{2,O,d}+\frac{k}{4b}\tau_{3,O,d}+\frac{1}{4}F_{T,d}\\[3pt]
    -\frac{k}{2b}\tau_{3,O,d}+\frac{1}{2}F_{T,d}\\[3pt]
    \frac{1}{F_{T,d}}\tau_{1,O,d}-\frac{k}{2bF_{T,d}}\tau_{3,O,d}+\frac{l}{2}
    \end{bmatrix}.
\end{equation}
The value of $d_{1,C}$, which is not used in Eq. (\ref{eq:c2foru}), remains zero through a separate controller.


In case of dual motor failure, we can configure $\mathbf{c}_{jk}\in\mathbb{R}^{4\times1}$ and $\mathbf{A}_{jk}(F_T)\in\mathbb{R}^{4\times4}$ by excluding terms and matrix columns related to the faulty motors $j$ and $k$ and utilizing both $d_{1,C}$ and $d_{2,C}$.
However, if the combination of failed motors is $\{j=1$, $k=3\}$ or $\{j=2$, $k=4\}$, $rank\left(\mathbf{A}_{jk}\right)$ becomes 3, which makes nominal 4-c-DOF flight impossible.
But otherwise, the signal $\mathbf{c}_{jk,d}$ can always be found to satisfy all components of $\mathbf{u}_d$ vector.

\subsection{Fail-safe Flight - Dynamics}
In the control process of a multirotor, the $\mathbf{u}_d$ signal generated by the flight computer is first converted into $\mathbf{c}_{k,d}$ through $\mathbf{A}_k^{-1}(F_T)$ in Eq. (\ref{eq:u_generation_failsafe}), and then passes the actuator dynamics to become $\mathbf{c}_k$ and $\mathbf{u}$.
In a conventional multirotor flight (refer Eq. (\ref{eq:u_conven})), the relationship of $\mathbf{u}_d=\mathbf{u}$ holds since the effects of rotor dynamics are negligible \cite{drone_basics}.

However, in the case of the $T^3$-Multirotor, not only the thrusters but also the servomechanism are implemented to generate $\mathbf{T}_O$ value and therefore, the assumption of $\mathbf{u}_d=\mathbf{u}$ is no longer valid.
Taking fail-safe flight for Motor 2 failure as an example, we can see from Eq. (\ref{eq:uM2}) that not only $F_4$ but also $d_{2,C}$ contribute to roll torque generation.
Since the relocation of the CoG position is accompanied by a change in the relative attitude $\alpha_{\{r,p\}}$ (refer Eq. (\ref{eq:X_CoG_to_rel_att})), there is a non-negligible dynamic relationship between $\tau_{1,O,d}$ and $\tau_{1,O}$.
Meanwhile, the remaining $\mathbf{u}$ terms do not suffer from the internal dynamics since those terms are still generated only through the motor thrusters with negligible rotor dynamics (refer Eq. (\ref{eq:uM2})).
{Therefore, in this subsection, we investigate only the EoM of $\alpha_{\{r,p\}}$-induced torque channel, by exploring the relationship between $\tau_{1,O,d}$ and $\tau_{1,O}$ in the Motor 2 failure scenario (\textit{Mode} 1).
Due to the symmetry of the $T^3$-Multirotor structure, the dynamics between $\tau_{r,O,d}$ and $\tau_{r,O}$ derived in Motor 2 failure scenario can also be applied to explain the behavior of other $\alpha_{\{r,p\}}$-induced torque control channels in different fail-safe flight scenarios.

Prior to investigation, we introduce some constraints during fail-safe flight for stable operation.
\begin{itemize}
    \item Constraint 1: Of the components of $\mathbf{u}$, $\tau_{3,O}$ is controlled to remain small during the fail-safe flight.
    \item Constraint 2: Of the components of $\mathbf{u}$, $F_{T}$ is controlled to remain near $Mg$ during the fail-safe flight.
\end{itemize}
By applying the Constraint 1 to Eq. (\ref{eq:uM2}), we can obtain the following relationship:
\begin{equation}
    \label{eq:F_constraints}
    F_1+F_3\approx F_4.
\end{equation}
Then, applying Eq. (\ref{eq:F_constraints}) to the Constraint 2 yields the following result:
\begin{equation}
    \label{eq:F4_constraint}
    F_4\approx\frac{1}{2}Mg.
\end{equation}
From Eqs. (\ref{eq:uM2}) and (\ref{eq:F4_constraint}), we can have the following equation:
\begin{equation}
    \label{eq:d_idle}
    \tau_{1,O}\approx Mg\left(d_{2,C}-d_{2,C}^{idle}\right)\ d_{2,C}^{idle}=\frac{l}{2}.
\end{equation}
Eq. (\ref{eq:d_idle}) shows that the value of $d_{2,C}$ should be biased near $d_{2,C}^{idle}$ to make $\tau_{1,O}$ near zero.
As a result, $\alpha_{rs}$ and $\tau_{rs}$ also should be biased during fail-safe flight.
By applying Eq. (\ref{eq:d_idle}) to (\ref{eq:rot_EoM_simplified}), (\ref{eq:T_CT_T_CF_nominal}) and (\ref{eq:X_CoG_to_rel_att}), we can obtain the following representation of $\alpha_r$ and $\tau_{rs}$.
\begin{equation}
    \label{eq:idles}
    \left\{
    \begin{aligned}
    \alpha_r&=\alpha_r^{idle}+\Delta_{\alpha_r},\ \alpha_r^{idle}\approx\frac{lM}{2m_Fd_F}\\
    \tau_{rs}&=\tau_{rs}^{idle}+\Delta_{\tau_{rs}},\ \tau_{rs}^{idle}=\frac{Mgl}{2}
    \end{aligned}
    \right.
\end{equation}
where small-angle approximation (SMA) is applied during $\alpha_r$ derivation.


\begin{figure}[t]
    \begin{center}
    \includegraphics[width=1\columnwidth]{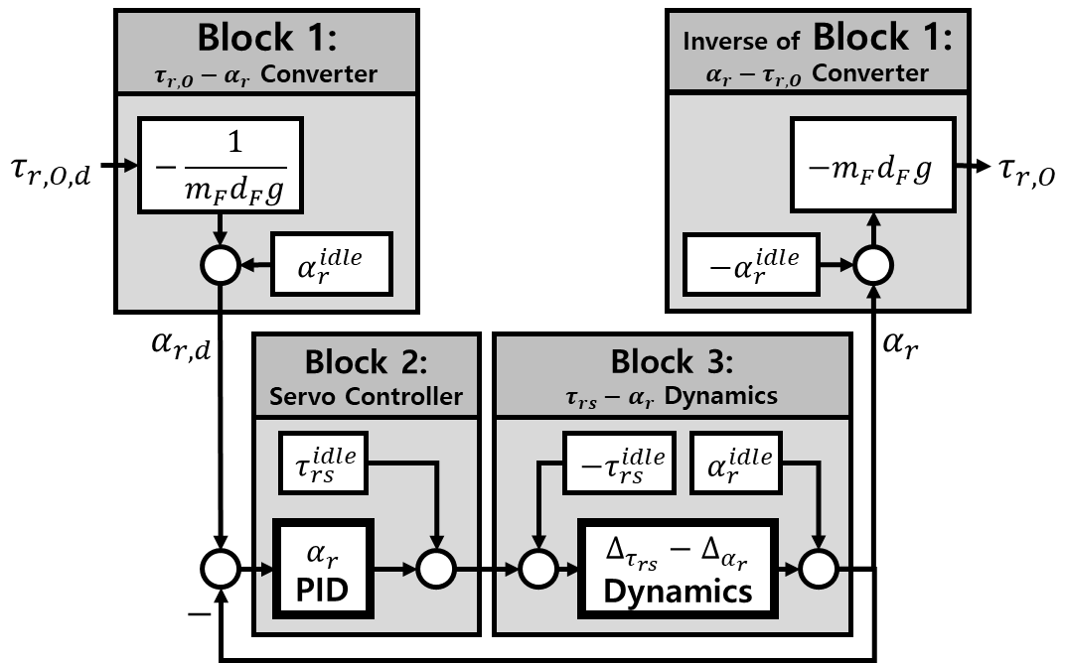}
    \end{center}
    \vspace{-0.5cm}
    \caption{Relationship between $\tau_{1,O,d}$ and $\tau_{1,O}$ in \textit{Mode} 1 fail-safe flight.}
    \vspace{-0.5cm}
    \label{gp:dynamics_failsafe}
\end{figure}

Fig. \ref{gp:dynamics_failsafe} shows the overall torque generation process.
The process can be divided into three blocks, Block 1, 2 and 3.

\subsubsection{Block 1}
Block 1 relates to the process of converting $\tau_{1,O,d}$ to $\alpha_{r,d}$ and reverting $\alpha_r$ to $\tau_{1,O}$.
For conversion, the $\tau_{1,O,d}$ signal is first converted to $d_{2,C,d}$ via Eq. (\ref{eq:d_idle}) and then converted to $\alpha_{r,d}$ through Eq. (\ref{eq:X_CoG_to_rel_att}).
During the conversion process, $\alpha_p$ is kept at zero value.
By combining Eqs. (\ref{eq:X_CoG_to_rel_att}) and (\ref{eq:d_idle}), we can obtain the following conversion equation between $\tau_{1,O}$ and $\alpha_r$.
\begin{equation}
    \label{eq:Block1}
    \alpha_r=\frac{1}{m_Fd_Fg}\tau_{1,O}+\alpha_r^{idle}
\end{equation}
In the same way, $\alpha_r$ is reverted to $\tau_{1,O}$ and utilized for TP attitude control.

\subsubsection{Block 2}
Block 2 is a controller block for tracking $\alpha_{r,d}$ generated in Block 1, where we can apply a simple linear controller (e.g. PID controller).
However, to overcome the bias of $\tau_{rs}$ identified in Eq. (\ref{eq:idles}) (which is $\tau_{rs}^{idle}$), we first generate the $\Delta_{\tau_{rs},d}$ signal through a `Servo Ctrler' block and then merge the $\tau_{rs}^{idle}$ value for generating $\tau_{rs,d}$ signal.

\subsubsection{Block 3}
Block 3 relates to the EoM between servo torque $\tau_{rs}$ and relative attitude $\alpha_r$.
From Eq. (\ref{eq:rot_EoM_simplified}), we can bring the rotational dynamics of the roll channel of TP and FP as follows.
\begin{equation}
    \label{eq:roll_dynamics}
    \left\{
    \begin{array}{lr}
    J_{1,T}\ddot{q}_{1,T}=\tau_{1,T}+\tau_{rs}\\
    J_{1,F}\ddot{q}_{1,F}=\frac{m_Fd_F}{M}F_T\sin{\alpha_r}\cos{\alpha_p}-\tau_{rs}
    \end{array}
    \right.
\end{equation}
When $F_2=0$, $F_4$ has a nearly fixed value by Eq. (\ref{eq:F4_constraint}).
Thus, applying Eq. (\ref{eq:F4_constraint}) to Eq. (\ref{eq:u_conven}) brings a nearly fixed $\tau_{1,T}$ value of
\begin{equation}
    \label{eq:fixed_tau_rt}
    \tau_{1,T}\approx-\frac{lMg}{2}
\end{equation}
during fail-safe flight.
Then, since $\alpha_r=q_{1,T}-q_{1,F}$ and $\alpha_p=0$, we can have the following equation from Eq. (\ref{eq:roll_dynamics}):
\begin{equation}
    \label{eq:alpha_dynamics}
    \ddot{\alpha}_r-\left(\frac{m_Fd_Fg}{J_{1,F}}\right)\alpha_r=\left(\frac{J_{1,T}+J_{1,F}}{J_{1,T}J_{1,F}}\right)\tau_{rs}-\frac{lMg}{2J_{1,T}},
\end{equation}
where aforementioned fail-safe flight constraints ($\tau_{3,O}\approx0$, $F_{T}\approx Mg$) and SMA are applied during the development.
Next, by applying Eq. (\ref{eq:idles}) to (\ref{eq:alpha_dynamics}), we obtain the following linear input/output relationship
\begin{equation}
    \label{eq:delta_relationship}
    \ddot{\Delta}_{\alpha_r}-\left(\frac{m_Fd_Fg}{J_{1,F}}\right)\Delta_{\alpha_r}=\left(\frac{J_{1,T}+J_{1,F}}{J_{1,T}J_{1,F}}\right)\Delta_{\tau_{rs}},
\end{equation}
which corresponds to the `$\Delta_{\tau_{rs}}-\Delta_{\alpha_r}$ Dynamics' block in Fig. \ref{gp:dynamics_failsafe}.
From Equations (\ref{eq:idles}) and (\ref{eq:delta_relationship}), we can express the relationship between $\tau_{rs}$ and $\alpha_r$ as shown in Block 3.

In Fig. \ref{gp:dynamics_failsafe}, we can see that all the blocks except the `Servo Ctrler' and `$\Delta_{\tau_{rs}}-\Delta_{\alpha_r}$ Dynamics' blocks cancel out.
Therefore, we can obtain the final form of fail-safe torque transfer function
\begin{equation}
    \label{eq:tf_roll_torque}
    \Lambda_{FS}(s)=\frac{\tau_{1,O}(s)}{\tau_{1,O,d}(s)}=\frac{\Lambda_{SC}(s)\Lambda_{ID}(s)}{1+\Lambda_{SC}(s)\Lambda_{ID}(s)},
\end{equation}
where
\begin{equation}
    \label{eq:transferfunction_torque_relatti}
    \Lambda_{ID}(s)=\frac{J_{1,T}+J_{1,F}}{J_{1,T}J_{1,F}s^2-m_Fd_Fg}
\end{equation}
is the transfer function of `$\Delta_{\tau_{rs}}-\Delta_{\alpha_r}$' block correspond to Eq. (\ref{eq:delta_relationship}), and $\Lambda_{SC}(s)$ is a transfer function correspond to the `Servo Ctrler' block.

Based on the above results, we can now describe the input-output relationship between $\mathbf{u}_d$ and $\mathbf{u}$ as follows:
\begin{equation}
    \label{eq:udu}
    \begin{split}
    &\mathbf{\Lambda}_\mathbf{u}(s)=\frac{\mathbf{u}(s)}{\mathbf{u}_d(s)}=diag\left(\mathbf{\Lambda}_{t}(s),\Lambda_{f}(s)\right),\\
    &\mathbf{\Lambda}_{t}(s)=diag\left(\Lambda_{t1}(s),\Lambda_{t2}(s),\Lambda_{t3}(s)\right),\\
    &\Lambda_{ti}(s)=\frac{T_{i,O}(s)}{T_{i,O,d}(s)}=\Lambda_{FS}(s)
    \end{split}
\end{equation}
where $\Lambda_{t\{1,2,3\}}(s),\Lambda_{f}(s)=1$ except for the $\Lambda_{ti}(s)$, which becomes $\Lambda_{FS}(s)$ when the \textit{Mode} $i$ fail-safe flight algorithm is activated.

\section{CONTROLLER DESIGN}

In this section we introduce a fail-safe flight controller design.
First, we introduce the faulty motor detection method, along with the proposed fail-safe flight controller.

\subsection{Faulty Motor Detection}

Before the motor failure, both $d_{\{1,2\},C}$ values are all fixed to zero.
Then, the attitude control torque is generated only by $\mathbf{T}_T$ according to the rules in Eq. (\ref{eq:u_conven}).
In this case, the rotational motion equation is derived from Eq. (\ref{eq:rot_EoM_finall}) and becomes as follows.
\begin{equation}
    \label{eq:general_multirotor_rot_EoM}
    \left(\mathbf{J}_T+\mathbf{J}_F\right)\ddot{\mathbf{q}}_T=\mathbf{T}_T,\ \ddot{\mathbf{q}}_F=\ddot{\mathbf{q}}_T
\end{equation}
Also, under normal flight conditions, $\mathbf{T}_T$ of $\mathbf{u}$ in Eq. (\ref{eq:u_conven}) has a limited range of values, and $F_T$ has a value near $Mg$ to overcome gravity, so roughly, each $F_i$ maintains a value near $0.25Mg$.
However, in the event of a single motor failure, the platform can no longer compensate the torque generated by the motor opposite the failed motor, resulting in an abnormally large roll or pitch torque with a magnitude of about $0.25lMg$.
From this fact, we can specify a faulty motor by monitoring whether the angular acceleration values $\ddot{q}_{1,T}$ or $\ddot{q}_{2,T}$ measured from IMU exceeds the conditions in Table I.
The value $\beta\in\mathbb{R}$ is derived from Eq. (\ref{eq:general_multirotor_rot_EoM}) and defined as 
\begin{equation}
    \label{eq:beta}
    \beta=\gamma\frac{0.25lMg}{J_{\{1,2\},T}+J_{\{1,2\},F}},
\end{equation}
where the value $\gamma\in\mathbb{R}$ is a heuristically adjustable parameter.

\begin{table}[b]
\vspace{-0.5cm}
\caption{FAULTY MOTOR IDENTIFICATION TABLE}
\label{tb:failure_id}
    \centering
    \vspace{-0.3cm}
    \begin{threeparttable}[t]
    \centering
        \begin{tabular}{lrlr}
        \toprule 
        Condition               & Faulty Motor      & Condition                 & Faulty Motor      \\ 
        \midrule
        $\ddot{q}_{1,T}>\beta$  & Motor 4 ($k=4$)   & $\ddot{q}_{2,T}>\beta$    & Motor 3 ($k=3$)   \\
        $\ddot{q}_{1,T}<-\beta$ & Motor 2 ($k=2$)   & $\ddot{q}_{2,T}<-\beta$   & Motor 1 ($k=1$)   \\
        \bottomrule
        \end{tabular}
    \end{threeparttable}
\end{table}

\subsection{Controller Design}

\begin{figure}[t]
    \begin{center}
    \includegraphics[width=1\columnwidth]{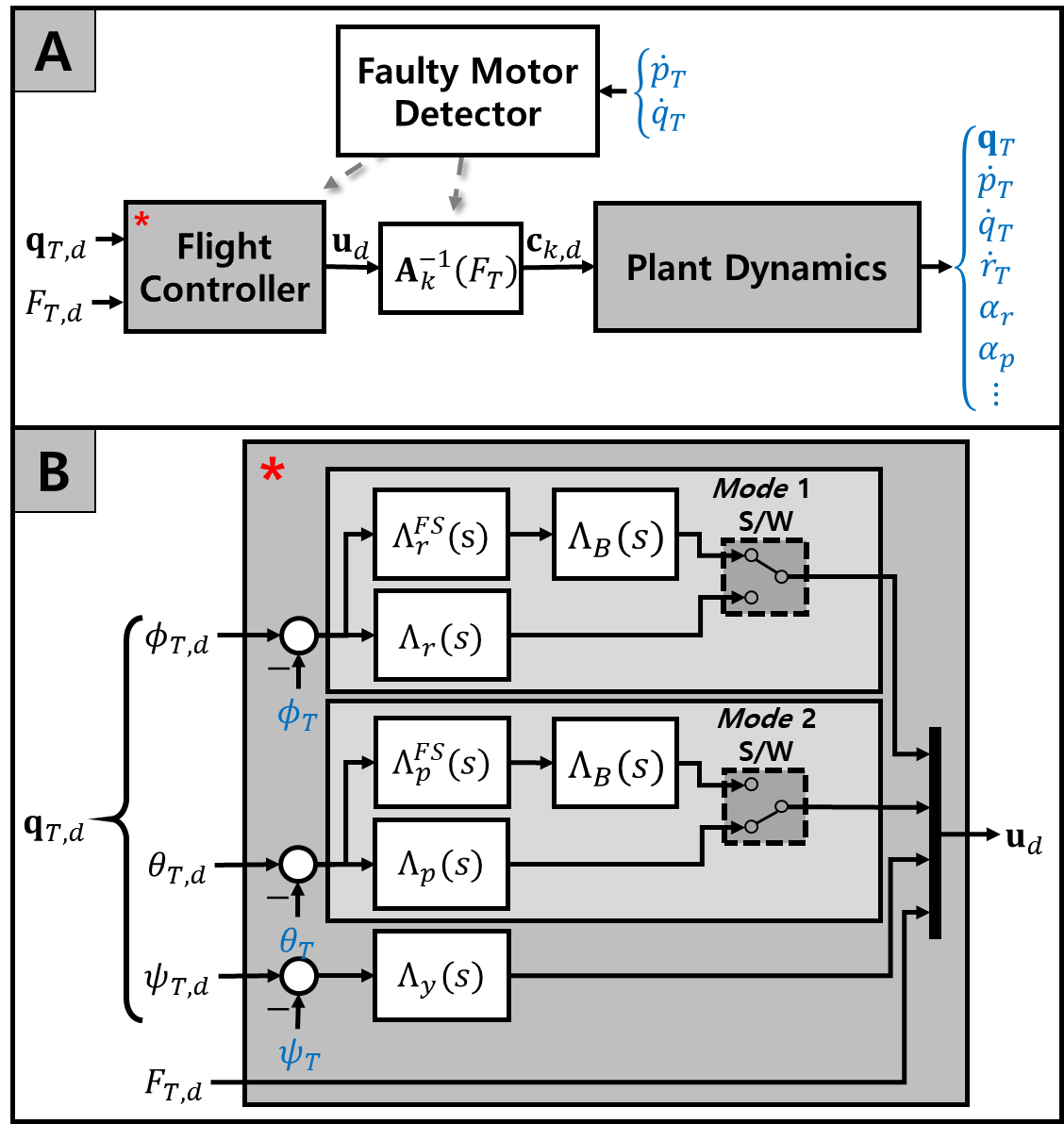}
    \end{center}
    \vspace{-0.5cm}
    \caption{The control scheme of the $T^3$-Multirotor with a fail-safe algorithm.}
    \vspace{-0.5cm}
    \label{gp:controller}
\end{figure}

Fig. \ref{gp:controller}-A shows the overall structure of the fail-safe flight system designed to achieve the proposed fail-safe flight.
The system consists of four blocks: `Faulty Motor Detector (FMD)', `Flight Controller (FC)', `Signal Converter (SC)', and `Plant Dynamics (PD)'.

\subsubsection{FMD Block}
The Faulty Motor Detector identifies the faulty motor $k$ from Table \ref{tb:failure_id}.
The information of the faulty motor is then sent to the `Flight Controller' block and 'Signal Converter' block to activate \textit{Mode} $i$ fail-safe flight algorithm.

\subsubsection{FC Block}
Fig. \ref{gp:controller}-B illustrates the internal structure of the Flight Controller Block.
The role of this block is to convert $\mathbf{q}_{T,d}$ and $F_{T,d}$ into $\mathbf{u}_d$, where those signals are generated by the high-level controller (e.g. position controller) or by the R/C controller.
During normal flight, dedicated roll, pitch and yaw feedback controllers ($\Lambda_{\{r,p,y\}}$), which ensure control stability and performance for the rotational motion of Eq. (\ref{eq:general_multirotor_rot_EoM}), generate $\mathbf{T}_{T,d}$(=$\mathbf{T}_T$) to control $\mathbf{q}_T$.
However, in the event of a motor failure, the roll and pitch dynamics changes and therefore, dedicated fail-safe controllers ($\Lambda_{\{r,p\}}^{FS}$) are configured to altered dynamics.
Each roll and pitch controller has a \textit{Mode} $i$ switch as in Fig. \ref{gp:controller}-B, where the FMD triggers the `\textit{Mode} 1 S/W' when $i=1$, and the `\textit{Mode} 2 S/W' when $i=2$.  

As we see in Eq. (\ref{eq:d_idle}), in the fail-safe flight mode, the servomotor is the only actuator for controlling the crippled attitude channel.
However, the servo motor has significantly slow response compared to the motor thruster due to their inherent characteristics, which may causes destructive vibration when high-frequency control input is applied.
To avoid this issue, a low pass filter ($\Lambda_B(s)$) is introduced to limit the frequency of the control input applied to the servo motor. 
In our research, a Bessel filter with the maximally flat phase delay characteristics is introduced to prevent an additional phase delay.

\subsubsection{SC and PD blocks}
For the SC block, the matrix is first set to $\mathbf{A}_0^{-1}$ in normal flight.
But in fail-safe flight, the FMD triggers the block to change the matrix to $\mathbf{A}_k^{-1}(F_T)$ for generating fail-safe flight command $\mathbf{c}_{k,d}$.

For the PD block, the system characteristics differs in normal and fail-safe flight.
In normal flight, the transfer function between $\mathbf{T}_{T,d}$ (of $\mathbf{u}_d$) and $\mathbf{q}_T$ is obtained from Eq. (\ref{eq:general_multirotor_rot_EoM}). 
We define the relationship as 
\begin{equation}
    \label{eq:transfer_Tq}
    \small
    \begin{split}
    &\mathbf{\Lambda}_{TA}^{N}(s)=\frac{\mathbf{q}_T(s)}{\mathbf{T}_{T}(s)}=diag(\Lambda_{1,TA}^{N}(s),\Lambda_{2,TA}^{N}(s),\Lambda_{3,TA}^{N}(s)),\\
    &\Lambda_{n,TA}^{N}(s)=\frac{1}{(J_{n,T}+J_{n,F})s^2},\ n=1,2,3,
    \end{split}
\end{equation}    
where $\mathbf{T}_{T,d}(s)=\mathbf{T}_T(s)$ due to negligible rotor dynamics.
However, in a fail-safe flight mode, Eq. (\ref{eq:rot_EoM_final}) governs the $\mathbf{q}_T$ motion and therefore, a new relationship of $\mathbf{\Lambda}_{TA}^{FS}(s)$ is defined as follows.
\begin{equation}
    \label{eq:uqt_FS}
    \small
    \begin{split}
    &\mathbf{\Lambda}_{TA}^{FS}(s)=\left\{
    \begin{array}{lr}
    diag(\Lambda_{1,TA}^{FS}(s)\ \Lambda_{2,TA}^{N}(s)\ \Lambda_{3,TA}^{N}(s))\ (if\ i=1)\\[3pt]
    diag(\Lambda_{1,TA}^{N}(s)\ \Lambda_{2,TA}^{FS}(s)\ \Lambda_{3,TA}^{N}(s))\ (if\ i=2)
    \end{array}
    \right.,\\
    &\Lambda_{i,TA}^{FS}(s)=\frac{q_{i,T}(s)}{T_{i,O}(s)}\Lambda_{ti}(s)=\frac{1}{J_{i,T}s^2}\Lambda_{FS}(s)
    \end{split}
\end{equation}

Then, the input-output relationship between $\mathbf{I}_{X,d}$ and $\mathbf{I}_X$ can be written as follows:
\begin{equation}
    \label{eq:Ixd_Ix}
    \begin{split}
    &\mathbf{\Lambda}_I(s)=\frac{\mathbf{I}_X(s)}{\mathbf{I}_{X,d}(s)}=diag(\Lambda_{1,I}(s),\Lambda_{2,I}(s),\Lambda_{3,I}(s),\Lambda_{4,I}(s)),\\
    &\Lambda_{1,I}(s)=\left\{
    \begin{array}{lr}
    \frac{\Lambda_r(s)\Lambda_{1,TA}^{N}(s)}{1+\Lambda_r(s)\Lambda_{1,TA}^{N}(s)}\ (if\ i\neq1)\\[5pt]
    \frac{\Lambda_r^{FS}(s)\Lambda_{1,TA}^{FS}(s)}{1+\Lambda_r^{FS}(s)\Lambda_{1,TA}^{FS}(s)} (if\ i=1)\\
    \end{array}
    \right.,\\
    &\Lambda_{2,I}(s)=\left\{
    \begin{array}{lr}
    \frac{\Lambda_p(s)\Lambda_{2,TA}^{N}(s)}{1+\Lambda_p(s)\Lambda_{2,TA}^{N}(s)}\ (if\ i\neq2)\\[5pt]
    \frac{\Lambda_p^{FS}(s)\Lambda_{2,TA}^{FS}(s)}{1+\Lambda_p^{FS}(s)\Lambda_{2,TA}^{FS}(s)} (if\ i=2)\\
    \end{array}
    \right.,\\
    &\Lambda_{3,I}(s)=\frac{\Lambda_y(s)\Lambda_{3,TA}^{N}(s)}{1+\Lambda_y(s)\Lambda_{3,TA}^{N}(s)},\ \Lambda_{4,I}(s)=\Lambda_{4,TA}^{N}(s),
    \end{split}
\end{equation}


\section{EXPERIMENT RESULT}

\subsection{Experimental Settings}

\begin{table}[b]
\vspace{-0.5cm}
\caption{PHYSICAL QUANTITIES AND CONTROLLER GAINS OF THE EXPERIMENTAL PLATFORM}
\label{tb:parameters}
    \vspace{-0.3cm}
    \centering
    \begin{threeparttable}[t]
    \centering
    \begin{tabular}{lrlr}
        \toprule 
        \multicolumn{4}{c}{Physical Parameters}                                                                                                         \\
        \toprule 
        Parameter   & Value                 & Parameter         & Value                                                                                 \\
        \midrule
        $m_T$       & \SI{0.389}{\kilogram} & $J_{\{1,2\},T}$   & $\approx$ \SI[inter-unit-product=\ensuremath{\cdot}]{0.002}{\kilogram\meter\squared}  \\
        $m_F$       & \SI{0.921}{\kilogram} & $J_{3,T}$         & $\approx$ \SI[inter-unit-product=\ensuremath{\cdot}]{0.01}{\kilogram\meter\squared}   \\
        $d_T$       & \SI{0.02}{\meter}     & $J_{\{1,2\},F}$   & $\approx$ \SI[inter-unit-product=\ensuremath{\cdot}]{0.014}{\kilogram\meter\squared}  \\ 
        $d_F$       & \SI{0.21}{\meter}     & $J_{3,F}$         & $\approx$ \SI[inter-unit-product=\ensuremath{\cdot}]{0.04}{\kilogram\meter\squared}   \\ 
        $l$         & \SI{0.15}{\meter}     & $b/k$ ratio       & $\approx$ 0.05                                                                        \\ 
        $F_{max}$   & \SI{12.9}{\newton}    & \multicolumn{2}{c}{-}                                                                                     \\
        \bottomrule\\
        \toprule 
        \multicolumn{4}{c}{Controller Gains}                                                                                                                                            \\
        \toprule
        Gain                                & Value                                             & Gain                                      & Value                                     \\ 
        \midrule
        $\Lambda_{\{r,p\}}(s)$ PID   & 3, 0.5, 0.3                                       & $\Lambda_{\{r,p\}}^{FS}(s)$ PID    & 0.1, 0.1, 0.24                            \\ 
        $\Lambda_{SC}(s)$ PID         & 5, 0.1, 3                                         & $\Lambda_y(s)$ PID                & 0.3, 0.01, 0.06                           \\
        Pos. PID                   & 2, 0.5, 2                                         & Height PID                         & 10,1,1                                    \\
        $\Lambda_B$ cut-off freq.         & \SI{40}{\hertz}                                   & \multicolumn{2}{c}{-}                                                                 \\
\bottomrule
    \end{tabular}
    \end{threeparttable}
\end{table}

The `$T^3$-Inverted' platform in Fig. \ref{gp:t3_inverted} is selected in the experiment to prevent airflow obstruction of the thrusters by the FP during relative attitude control.\footnote{See {https://www.sjlazza.com/post/ra-l-2020} for hardware requirements for fail-safe flight.}
Table II shows the hardware parameters and the controller gains of the experimental platform.
For controllers, PID controllers are introduced in all of $\Lambda_{SC}(s)$, $\Lambda_{\{r,p,y\}}(s)$ and $\Lambda_{\{r,p\}}^{FS}(s)$, and the gains are set to make $\mathbf{\Lambda}_I(s)$ a stable system.
The $\mathbf{I}_{X,d}$ command is generated by the dedicated PID position controller.
However, the Bessel filter is limited to have a cutoff frequency of \SI{40}{\hertz} due to the inherent characteristics of the servomotor.



\begin{figure}[t]
    \begin{center}
    \includegraphics[width=1\columnwidth]{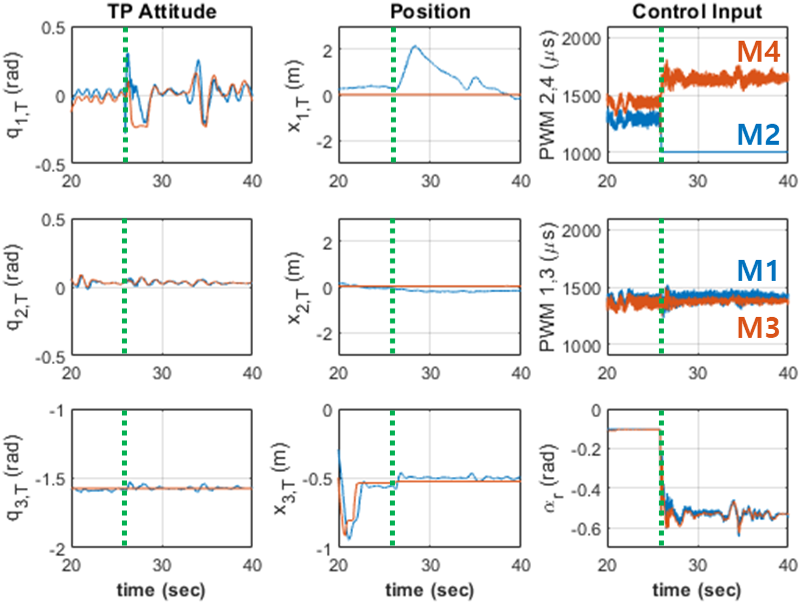}
    \end{center}
    \vspace{-0.5cm}
    \caption{\textbf{[Fail-safe flight \#1]} \textbf{[}\textcolor{red}{Red}: ref. trajectory, \textcolor{blue}{Blue}: tracking result\textbf{]} Attitude and position tracking results before and after motor failure. Motor 2 failure occurred at around 26 seconds, which triggered \textit{Mode} 1 fail-safe flight controller.}
    \vspace{-0.5cm}
    \label{gp:flight}
\end{figure}

\subsection{Flight Results}
In the experiment, the motor failure is caused by the operator triggering the stop signal of Motor 2 at any time.
Once the motor fails, the FMD identifies the failed motor and activates the fail-safe roll control mode, whilst the pitch channel remains the conventional control mode.

Fig. \ref{gp:flight} shows the fail-safe flight results.
The figure includes the tracking results of the TP attitude, the three-dimensional platform position, the four propeller PWM control inputs, and the relative roll attitude.
From the PWM 2 command log in the `Control Input' graph, we can see that Motor 2 stopped rotating at about 26 seconds.
After that, the FMD triggers the \textit{Mode} 1 fail-safe flight to control the servo motor for restoring the crippled roll attitude.
As a result, not only the roll attitude control but also the position control was successfully restored, and the $x$-direction position converged to the target position within 10 seconds after the event. 

Fig. \ref{gp:flight2} shows the tracking results when changing $q_{3,T,d}$, $x_{1,T,d}$ and $x_{3,T,d}$ values to validate heading, horizontal and vertical motion control performance during fail-safe flight.
Experimental results show satisfactory heading and height control performance.
In the $x_{1,T}$ position control, however, the reference trajectory tracking performance is reduced compared to other channels.
This is mainly due to the limited cutoff frequency of the Bessel filter, which degrades the $\Lambda_{1,TA}^{FS}(s)$ performance, and the improvement of control performance via the advanced control method, such as the robust control method, remains a topic for future research.

\section{CONCLUSION}

In this paper, we introduced a new quadcopter fail-safe flight that utilizes a fully-actuated $T^3$ mechanism to maintain all four c-DOF control capabilities in the event of a single motor failure.
The proposed method applied the strategy of actively adjusting the CoG position of the fuselage to restore the c-DOF of the system.
A method for identifying faulty motors using only accelerometer information is devised, and a flight control system specialized for fail-safe flight to isolate faulty motors has been constructed.
To understand the motion characteristics during fail-safe flight, equations of motions in fail-safe flight are derived, and experimenal results are proved to validate that the independent control of roll, pitch, yaw and thrust is recovered by the proposed method after a failure of single thruster.

\begin{figure}[t]
    \begin{center}
    \includegraphics[width=1\columnwidth]{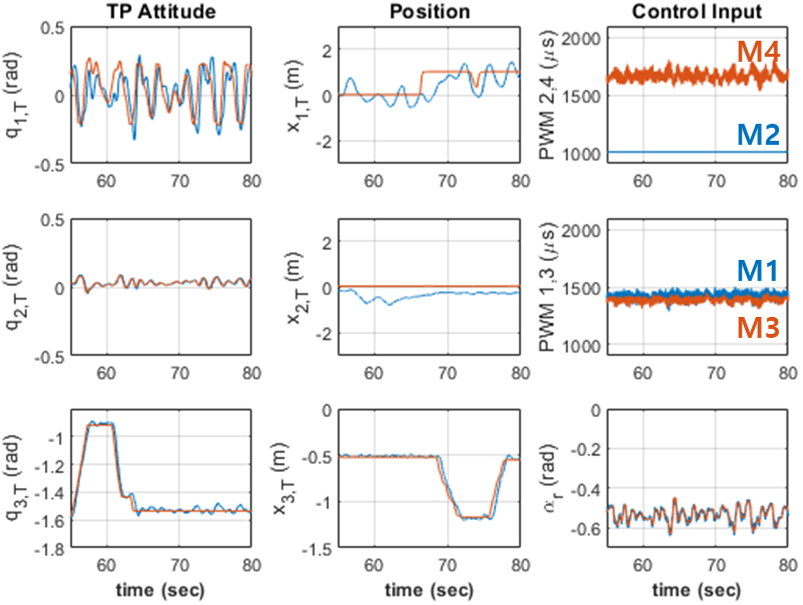}
    \end{center}
    \vspace{-0.5cm}
    \caption{\textbf{[Fail-safe flight \#2]} \textbf{[}\textcolor{red}{Red}: ref. trajectory, \textcolor{blue}{Blue}: tracking result\textbf{]} Reference trajectory tracking results when arbitrary $q_{3,T,d}$, $x_{1,T,d}$, and $x_{3,T,d}$ commands are applied.}
    \vspace{-0.23cm}
    \label{gp:flight2}
\end{figure}

This algorithm is advantageous in situations where the system requires enhanced operation safety but also needs to suppress the additional hardware, such as parachutes or additional thrusters that significantly increase the weight or volume of the platform.
It is also advantageous when the flight relies on the built-in navigation systems that are vulnerable to rotations, such as visual odometry cameras, as the proposed method can ensure a stable attitude during fail-safe flight.

\addtolength{\textheight}{-12cm}   









\end{document}